\documentclass[preprint,review,12pt,3p]{elsarticle}

\usepackage{amssymb}
\usepackage{amsmath}
\usepackage{graphicx}
\usepackage{multirow}
\usepackage{color}
\usepackage{subcaption}
\usepackage{caption}
\usepackage{float}
\usepackage[colorlinks,
            linkcolor=black,
            anchorcolor=black,
            citecolor=black
            ]{hyperref}

\journal{Pattern Recognition}

\begin{document}

\begin{frontmatter}

%% Title, authors and addresses

%% use the tnoteref command within \title for footnotes;
%% use the tnotetext command for theassociated footnote;
%% use the fnref command within \author or \affiliation for footnotes;
%% use the fntext command for theassociated footnote;
%% use the corref command within \author for corresponding author footnotes;
%% use the cortext command for theassociated footnote;
%% use the ead command for the email address,
%% and the form \ead[url] for the home page:
%% \title{Title\tnoteref{label1}}
%% \tnotetext[label1]{}
%% \author{Name\corref{cor1}\fnref{label2}}
%% \ead{email address}
%% \ead[url]{home page}
%% \fntext[label2]{}
%% \cortext[cor1]{}
%% \affiliation{organization={},
%%             addressline={},
%%             city={},
%%             postcode={},
%%             state={},
%%             country={}}
%% \fntext[label3]{}

\title{Versatile Teacher: A Class-aware Teacher-student Framework for Cross-domain Adaptation}

%% use optional labels to link authors explicitly to addresses:
%% \author[label1,label2]{}
%% \affiliation[label1]{organization={},
%%             addressline={},
%%             city={},
%%             postcode={},
%%             state={},
%%             country={}}
%%
%% \affiliation[label2]{organization={},
%%             addressline={},
%%             city={},
%%             postcode={},
%%             state={},
%%             country={}}

\author[1]{Runou Yang}
\ead{runouyang@hust.edu.cn}

\author[1]{Tian Tian \corref{cor1}}
\ead{ttian@hust.edu.cn}

\author[1]{Jinwen Tian}
\ead{jwtian@hust.edu.cn}

\address[1]{National Key Laboratory of Multispectral Information Intelligent Processing Technology, School of Artificial Intelligence and Automation, Huazhong University of Science and Technology, China}

\cortext[cor1]{Corresponding author}

%% Author affiliation
% \affiliation{organization={School of Artificial Intelligence and Automation, Huazhong University of Science and Technology},%Department and Organization
%             country={China}}

%% Abstract
\begin{abstract}
%% Text of abstract
Addressing the challenge of domain shift between datasets is vital in maintaining model performance. In the context of cross-domain object detection, the teacher-student framework, a widely-used semi-supervised model, has shown significant accuracy improvements. However, existing methods often overlook class differences, treating all classes equally, resulting in suboptimal results. Furthermore, the integration of instance-level alignment with a one-stage detector, essential due to the absence of a Region Proposal Network (RPN), remains unexplored in this framework. In response to these shortcomings, we introduce a novel teacher-student model named Versatile Teacher (VT). VT differs from previous works by considering class-specific detection difficulty and employing a two-step pseudo-label selection mechanism, referred to as Class-aware Pseudo-label Adaptive Selection (CAPS), to generate more reliable pseudo labels. These labels are leveraged as saliency matrices to guide the discriminator for targeted instance-level alignment. Our method demonstrates promising results on three benchmark datasets, and extends the alignment methods for widely-used one-stage detectors, presenting significant potential for practical applications. Code is available at \href{https://github.com/RicardooYoung/VersatileTeacher}{https://github.com/RicardooYoung/VersatileTeacher}.
\end{abstract}

%%Graphical abstract
% \begin{graphicalabstract}
%\includegraphics{grabs}
% \end{graphicalabstract}

%%Research highlights
% \begin{highlights}
% \item Research highlight 1
% \item Research highlight 2
% \end{highlights}

%% Keywords
\begin{keyword}
%% keywords here, in the form: keyword \sep keyword
Domain adaptation \sep Object detection \sep Mean teacher
%% PACS codes here, in the form: \PACS code \sep code

%% MSC codes here, in the form: \MSC code \sep code
%% or \MSC[2008] code \sep code (2000 is the default)

\end{keyword}

\end{frontmatter}
\biboptions{sort&compress}

%% Add \usepackage{lineno} before \begin{document} and uncomment 
%% following line to enable line numbers
%% \linenumbers

%% main text
%%

\section{Introduction}
\label{sec:intro}

When a well-trained neural network is presented with an unlabeled dataset that differs from its original training set, it is common for the model to suffer a degradation in accuracy. This phenomenon is called domain shift, wherein the training set and the test set are sampled from different distribution. In the context of object detection tasks, it will increase the difficulty of object recognition. Since labeling a new dataset is a labor-exhausted and expensive work, Unsupervised Domain Adaptation (UDA) emerges as a valuable tool to address this challenge. UDA aims to bridge the domain gap between a source domain (where the model is trained) and a target domain (the new, unlabeled dataset). This domain gap often arises due to factors like changing weather conditions, varying illumination and angles, or differences in the capturing device.

Researchers have proposed numerous UDA methods, broadly categorized into three streams: discrepancy-based methods\cite{changjianshuiNovelDomainAdaptation2022, chaochenHoMMHigherOrderMoment2020, mozhdehzandifarDomainAdaptationBregman2021}, reconstruction-based methods\cite{shih-chiahuangDSNetJointSemantic2020, 10.1007/978-3-031-26313-2_30}, and adver-sarial-based methods\cite{JMLR:v17:15-239, chen2018domain, shizhaozhangDomainAdaptiveYOLO2021, cheng-chunhsuEveryPixelMatters2020, mazinhnewaIntegratedMultiscaleDomain2023a, mazinhnewaMultiscaleDomainAdaptive2021}. These approaches leverage unlabeled data to extract domain-irrelevant features and improve the model's performance on the target domain. Among the three mentioned methods, adversarial-based methods have shown the most promising results. It is first proposed in domain-adversarial neural network (DANN)\cite{JMLR:v17:15-239}, which introduces image-level alignment. Furthermore, apart from that, instance-level alignment is proposed by employing adversarial learning on instance scale, resulting in improved performance. It has been proven that applying instance-level alignment can enhance the similarity of distributions across both domains, effectively overcoming domain mismatch\cite{chen2018domain}. However, performing instance-level alignment requires a Region Proposal Network (RPN) to generate potential locations of the objects, which is not included in one-stage algorithms such as YOLO\cite{josephredmonYouOnlyLook2016, josephredmonYOLOv3IncrementalImprovement2018, josephredmonYOLO9000BetterFaster2016}, making it challenging to be adopted for one-stage detectors.

In addition to the aforementioned methods, another approach involves employing a teacher-student structure. In this setup, the teacher model generates pseudo labels for the student model, transforming unsupervised learning into a semi-supervised learning (SSL) scenario. This type of method offers the advantage of not requiring any annotation on the target domain and can learn more intrinsic features compared to simple adversarial learning. One widely used framework is Mean Teacher (MT)\cite{anttitarvainenMeanTeachersAre2018}. Adaptive Teacher (AT)\cite{yu-jheliCrossDomainAdaptiveTeacher2022} takes MT as its pipeline, and incorporates image-level alignment and pseudo label regulation, resulting in accuracy gain. In AT, images from the target domain undergo weak and strong augmentations before being processed by the teacher and student models, respectively. Since pseudo labels generated by the teacher model may contain errors and noise, it is crucial to conduct screening. Nevertheless, AT does not adequately consider class differences and treat every class equally. Moreover, AT does not apply instance-level alignment to maintain compatibility with one-stage detectors, thereby limiting its effectiveness and application.

To address the problems mentioned above, a brand new teacher-student framework named Versatile Teacher (VT) is proposed in this paper, whose ability for domain adaptation is enhanced by generating more reliable pseudo labels. We design an novel selection procedure, called Class-aware Adaptively Pseudo-label Selection (CAPS). CAPS firstly filters out noise and low-confidence objects, and uses the rest to update a class-aware threshold for each class. Subsequently, the updated threshold is then employed for a second selection to obtain reliable pseudo labels. Furthermore, these pseudo labels are utilized as saliency matrices to denote where the objects are to perform targeted instance-level alignment, making location-aware alignment feasible for one-stage detectors, thereby broadening their potential applications in engineering.

In summary, our main contributions are threefold: 1) We propose a novel teacher-student framework that not only excels at bridging the domain gap but also stands as the first to integrate vanilla instance-level alignment with a one-stage detector, demonstrating a promising practical outlook. 2) We devise the CAPS mechanism, aimed at enhancing the reliability of pseudo labels by applying class-aware treatment. 3) We leverage pseudo labels as saliency matrices to guide targeted instance-level alignment for one-stage detectors, expanding the applicability of teacher-student framework in domain adaptation.

\section{Related Works}

\subsection{Object Detection}

The field of object detection encompasses two main types of algorithms: two-stage algorithms and one-stage algorithms. Two-stage algorithm follows the traditional object detection pipeline, generating region proposals firstly and then classifying each proposal into different object categories. The R-CNN series\cite{6909475, rossgirshickFastRCNN2015, 7485869, 8237584} is one of the most well-known structures of two-stage detectors. This type of algorithm typically consists of a feature extractor, a region proposal network (RPN), and a classifier. The feature extractor extracts features from the input image, the RPN generates potential object locations, and the classifier predicts whether each location contains an object and, if so, its corresponding class.

On the other hand, one-stage algorithms do not rely on an RPN and directly output prediction results from features. The YOLO series\cite{josephredmonYouOnlyLook2016, josephredmonYOLOv3IncrementalImprovement2018, josephredmonYOLO9000BetterFaster2016}, Single Shot MultiBox Detector (SSD)\cite{10.1007/978-3-319-46448-0_2} and RetinaNet\cite{Lin_2017_ICCV} are representative examples of one-stage detectors. While one-stage detectors may not achieve the same level of accuracy as two-stage detectors due to the absence of an RPN and a simpler feature extraction process, they offer significantly faster inference time, making them more suitable for real-world deployment scenarios.

\subsection{Unsupervised Domain Adaptation}

The adversarial-based domain adaptation method was first introduced in the Domain Adversarial Neural Network (DANN)\cite{JMLR:v17:15-239}. Inspired by the concept of Generative Adversarial Networks (GANs), DANN incorporates a domain discriminator into the network architecture, forging the adversarial interaction between feature extractor and discriminator. By doing so, the margin distributions of the two domains get closer, enabling the feature extractor to generalize better across both domains. \emph{Chen et al.}\cite{chen2018domain} further improved DANN by introducing an instance-level discriminator and a consistency regularization. The instance-level discriminator performs classification on the region proposals from RPN, instead of the whole image. This new branch enables the model to pay more attention on local differences such as appearance, size and perspective. Besides, a consensus regularization is raised to encourage both discriminators to generate consistent outputs. SCDA \cite{zhu2019adapting} utilizes a K-means clustering approach to identify discriminative regions across two domains, complemented by a GAN to achieve adjusted instance-level alignment. Diversify and Match (DM)\cite{kim2019diversify} employs CycleGAN to generate images from different domains, facilitating the learning of diverse domain shifts. Meanwhile, MeGA-CDA \cite{VS_2021_CVPR} enhances instance-level alignment by integrating category information. It utilizes a memory bank that stores features from various classes, allowing for the calculation of similarity between extracted features and those stored in the bank.

As for one-stage detectors, \emph{Hnewa et al.}\cite{mazinhnewaMultiscaleDomainAdaptive2021} proposed MS-DAYOLO, which combines image-level discriminators with YOLOv4. Building upon MS-DAYOLO, \emph{Hnewa et al.}\cite{mazinhnewaIntegratedMultiscaleDomain2023a} later introduced integrated MS-DAYOLO, which modified the approach to align image-level features. \emph{Zhang et al.}\cite{shizhaozhangDomainAdaptiveYOLO2021} introduced DA-YOLO, which incorporates instance-level alignment and consensus regularization. However, the instance-level alignment in DA-YOLO treats every position in the feature maps equally, rather than focusing on the objects, which goes against the original intention. \emph{Hsu et al.}\cite{cheng-chunhsuEveryPixelMatters2020} made further improvements to instance-level alignment by adding a small Fully Convolutional Network (FCN) to generate object probability map for each pixel in the feature maps. Nevertheless, this enhancement increases computational complexity, and a small FCN may not be capable to accurately predict the precise object positions. \emph{Krishna et al.}\cite{krishna2023mila} adopts a memory bank to store instance-level features from previous-seen source images. The instance-level alignment is performed between target and source features based on similarity.

\subsection{Mean Teacher}

Mean Teacher (MT) framework\cite{anttitarvainenMeanTeachersAre2018} was proposed to turn a unsupervised classification problem into a semi-supervised one. The framework contains a teacher model and a student model. In the process of handling labeled images, distinct augmentations are applied to the images before they are fed into the teacher and student models, respectively. Apart from classification loss, there is also a consistency loss to force two models generate similar features under different augmentations. The teacher model does not use backpropagation to update its weights, on the other hand it use student model weight to update via exponential moving average(EMA). 

Based on MT, MTOR\cite{cai2019exploring} introduces inter- and intra-graph consistency loss to match graph structures between teacher and student and enhance the similarity between regions of same class within the graph of student, respectively. CRDA\cite{xu2020exploring} integrates a multi-label classifier at the image level with the detection backbone to pinpoint sparse but crucial image regions related to categorical information. UMT\cite{deng2021unbiased} leverages pixel-level adaptation for images from both source and target domain to mitigate domain bias. \emph{Li et al.}\cite{yu-jheliCrossDomainAdaptiveTeacher2022} proposed Adaptive Teacher (AT) to further leverage the potential of the teacher-student structure, and transferred this methodology to detection task. It abandons the consistency loss, and uses the teacher model to generate pseudo labels directly, supervising detection mission of the student model. Moreover, it applies image-level domain discriminator to reduce domain gap. This framework was designed for all detectors, including one-stage detector, thus no instance-level alignment was adopted. Also, several studies\cite{kemeiInstanceAdaptiveSelftraining2020, inkyushinTwoPhasePseudoLabel2020, yangzouUnsupervisedDomainAdaptation2018} have shown that different class should be set different threshold to filter out credible pseudo labels, whereas AT set a universal threshold for all classes. CMT\cite{cao2023contrastive} is built upon AT, cooperating object-level constrasive loss, which is a variant of instance-level alignment, but without help of RPN. However, it necessitates a large amount of memory to store the features of the last several layers. SSDA-YOLO\cite{zhou2023ssda} shares the same architecture with MT, using a trained Contrastive Unpaired Translation (CUT)\cite{park2020contrastive} to generate extra pseudo training images first and calculating constrasive loss between real and fake images. The training of CUT makes the entire procedure verbose, and the results depend on the outcomes of CUT.

There are some exploration of adaptive thresholds in semi-supervised tasks to enhance the reliability of pseudo labels. PLCM\cite{Huang_2021_ICCV} evaluates the confidence of pseudo labels by measuring the entropy of the predicted distribution, which provides an indirect assessment of label certainty. FlexMatch\cite{NEURIPS2021_995693c1} dynamically adjusts thresholds according to the quantity of samples exceeding these thresholds, tailoring the difficulty of the task to the model’s current learning state. However, these advancements depend heavily on labeled data and are confined to the training dataset. CT\cite{Wang_2023_CVPR} utilizes a Gaussian Mixture Model (GMM) to assess the likelihood that the model accepts a pseudo-label as the ground truth under the assumption of Gaussian distribution prior. Being a parametric method, GMM’s ability of approximation is subject to the number of components. With the number of components increasing, the computational complexity of estimating parameters will grow rapidly.

\begin{figure*}
    \centering
    \includegraphics[width=0.9\textwidth]{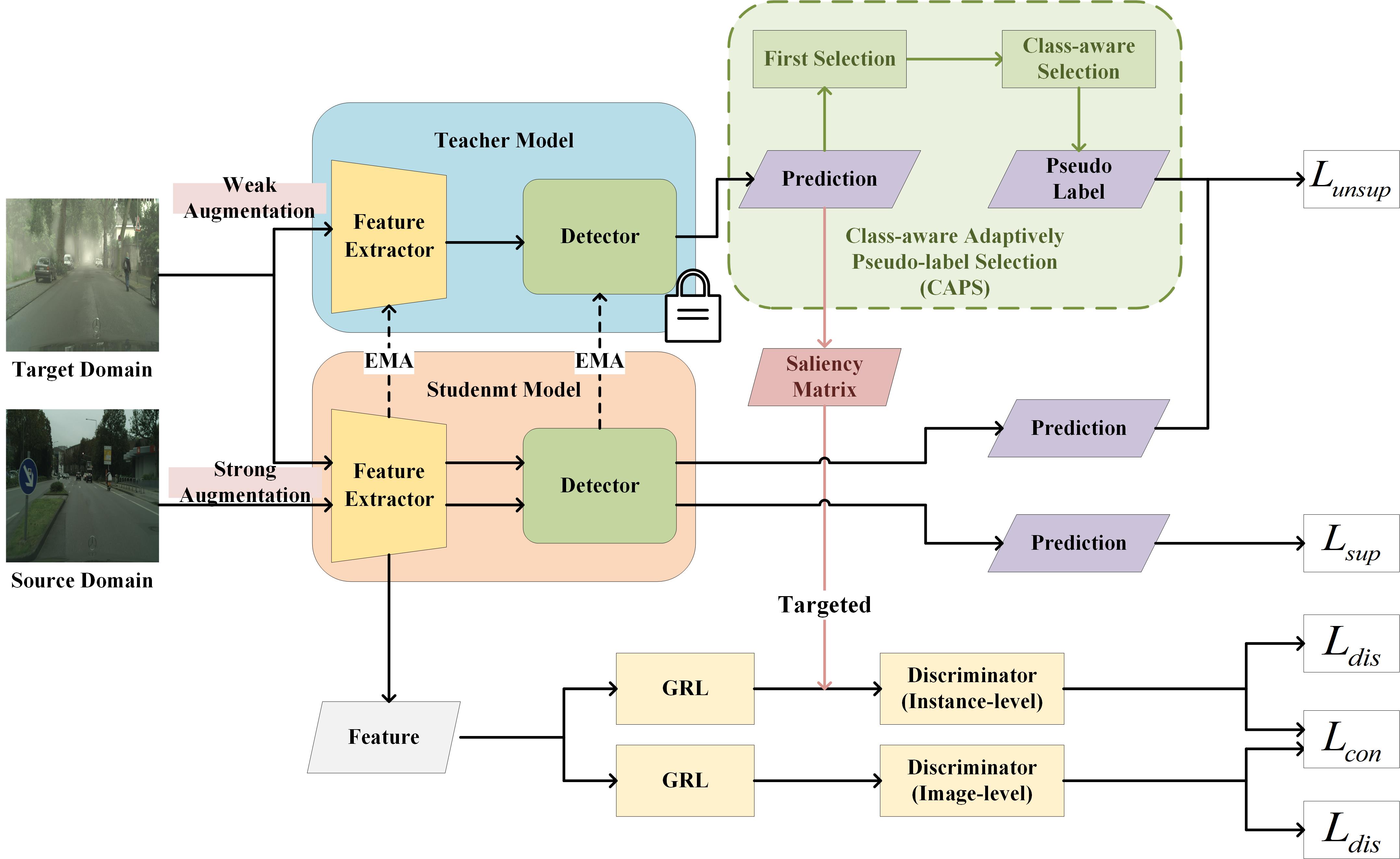}
    % \caption{The structure of Versatile Teacher framework. Input images are weakly augmented and fed to teacher model to generate pseudo labels. Student model receives images undergoing strong augmentation and compute loss with generated pseudo labels. Student model also contains a adversarial learning branch for image-level and instance-level alignment.}
    \caption{The structure of Versatile Teacher framework. Built upon AT, we design CAPS (the green part) to generate more reliable pseudo labels through a two-step selection process (detailed in Sec \ref{sec:mt}). We further leverage pseudo labels as saliency matrices (the red part) to perform targeted instance-level alignment (detailed in Sec \ref{sec:al}).}
    \label{fig:net}
\end{figure*}

\section{Proposed Method}

\subsection{Problem Definition and Overview}

Assuming that we have two distinct datasets in different domains: a training dataset $D^S=\{x^S_i, y^S_i,b_i^S\}_{i=1}^n$ with sufficient labeled data as source domain, where $x_i^S$, $y_i^S$, $b_i^S$ denote image, class label of the object, and its corresponding bounding box, respectively. And a test dataset, $D^T=\{x^T_j\}_{j=1}^m$, representing our target domain without any annotations. The aim of domain adaptation is to design a detector which can transfer knowledge learned from the source domain to the target domain and achieve higher accuracy in the target domain.

The proposed Versatile Teacher framework is visually depicted in Figure \ref{fig:net}. It consists of two pivotal models: the teacher model and the student model. The training process involves the following steps. Prior to training, labeled source data are used to get a pretrained model to initialize both models. During training, the teacher model receives images from the target domain $D^T$ after undergoing weak augmentation. In contrast, the student model receives images from both domains after undergoing strong augmentation. When processing images from the source domain, the student model computes the loss against their ground truth labels and updates its parameters accordingly. For images from the target domain, the teacher model first generates predictions. After a two-step adaptive selection process (detailed in Sec \ref{sec:mt}), pseudo labels and saliency matrices (detailed in Sec \ref{sec:al}) are derived. These pseudo labels serve as ground truth for the student model. How teacher model updates its parameters is not by backpropagation, but via EMA from student model's parameters. Our framework incorporates adversarial learning, featuring both image-level and instance-level domain discriminators. Following the approach by \emph{Chen et al.}\cite{chen2018domain}, we also introduce a consistency regularization term to enhance model performance (detailed in Sec \ref{sec:al}).

\subsection{Learning From Unlabeled Data}\label{sec:mt}

\subsubsection{Model Initialization}

In the teacher-student framework, the generation of reliable pseudo labels is crucial. We start by training our model on the source domain data, denoted as $\{x_i^S, y_i^S,b_i^S\}_{i=1}^n,$ using a supervised loss function $L_{sup}$. Taking YOLO as example, assuming that model's prediction on image $i$ is $(\hat{y}_i,\hat{b}_i)$, representing classification results and corresponding bounding boxes, the $L_{sup}$ in YOLO is defined as

\begin{equation}
    L_{sup}=\lambda_1L_{bbox}(\hat{b}_i, b_i^S)+\lambda_2L_{obj}(\hat{y}_i,y_i^S)+\lambda_3L_{cls}(\hat{y}_i,y_i^S),
\end{equation}

\noindent where $L_{bbox}$ is the bounding box loss, $L_{obj}$ is the confidence loss and $L_{cls}$ is the classification loss. $\lambda_1,\lambda_2,\lambda_3$ is their weight respectively.

Apart from that, data $\{x_j^T\}_{j=1}^m$ from target domain is also used to initialize domain discriminator using domain classification loss $L_{dis}$ and consensus loss $L_{con}$, which is detailed in Sec \ref{sec:al}. After pretraining is completed, parameters obtained are transferred to both teacher model and student model for initialization.

\subsubsection{Class-aware Adaptively Pseudo-label Selection (CAPS)}

In the absence of labels in the target domain, another way to offer supervision is using additional model to generate pseudo labels. But pseudo labels often contain noise that can negatively impact model performance. When converting predictions into pseudo labels directly, the risk of undermining the model's performance is high. To mitigate this, \emph{Li et al.}\cite{yu-jheliCrossDomainAdaptiveTeacher2022} employed a high threshold to filter out low-confidence predictions. However, different object classes may present varying detection difficulties, and applying a global threshold might overlook reliable predictions for some classes\cite{kemeiInstanceAdaptiveSelftraining2020, inkyushinTwoPhasePseudoLabel2020, yangzouUnsupervisedDomainAdaptation2018}. Setting a class-aware threshold for every class is a feasible solution to this problem. In CT\cite{Wang_2023_CVPR}, a simple 1-D Gaussian distribution is chosen to model the probability whether the network accepts the pseudo label, which have difficulty dealing with non-Gaussian. A novel non-parametric way named CAPS is proposed in our work, which aims to get the empirical distribution of confidence, offering better adaptability and robustness for various distributions. 

\begin{figure*}[!h]
    \centering
    \includegraphics[width=0.9\textwidth]{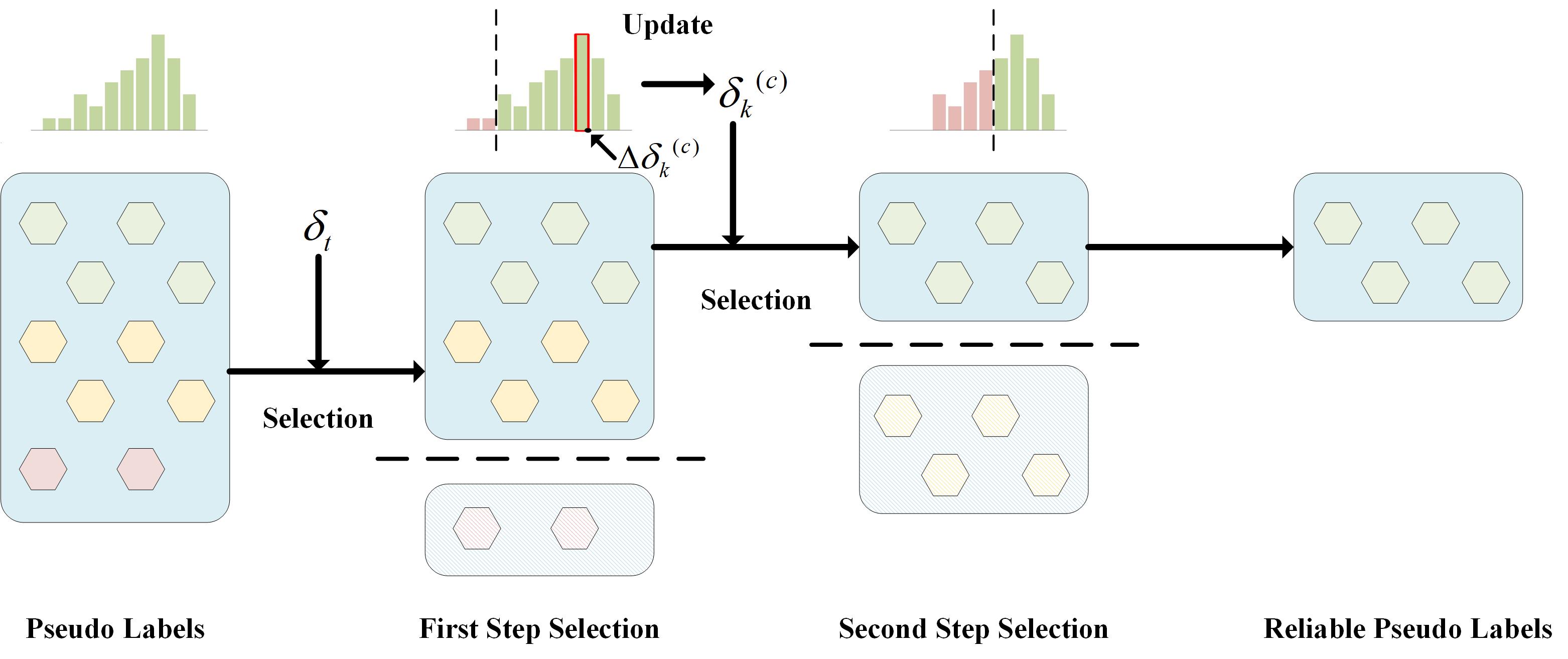}
    \caption{The procedure of CAPS. A low threshold will first filter out highly unreliable predictions. The remaining predictions update a high threshold with their mode of confidence. Afterwards the updated high threshold is used to select reliable pseudo labels.}
    \label{fig:as}
\end{figure*}

Before processing an image from the target domain, we apply both weak and strong augmentations. The weakly augmented version is passed to the teacher model to generate predictions. Then a two-step selection is performed on those results. The procedure is demonstrated in Figure \ref{fig:as}. First, a relatively low threshold $\delta_t$ is applied to filter out highly unreliable predictions across all classes. The confidences of remaining predictions are spread in $M$ intervals that split $[0,1]$ equally. Counting confidences falling on every interval per class can be roughly regarded as an approximation for the probability density function of distribution. We identify the interval with the highest confidence density, namely the peak of density curve, and set the right endpoint of this interval as $\Delta\delta^{(c)}$. For instance, when taking $M=20$, the length of every interval is 0.05. If most confidences fall into the interval $(0.8,0.85]$, i.e. the peak occurs here, $\Delta\delta^{(c)}$ will be set to 0.85 for class $c$. $\Delta\delta^{(c)}$ is used to update another threshold $\delta_k^{(c)}$ for class $c$ via EMA:

\begin{equation}
    \delta_k^{(c)}=\alpha_t\delta_{k-1}^{(c)}+(1-\alpha_t)\beta\Delta\delta^{(c)},
\end{equation}

\noindent in which $k$ is the number of iterations, $\beta$ serves as a factor in cosine annealing\cite{loshchilov2016sgdr} to achieve improved convergence..

$\delta_0^{(c)}$ is initialized with a rather high threshold. Afterwards a second selection using updated class-specific thresholds $\{\delta_k^{(c)}\}$ is adopted. For prediction belongs to class $c$ with a confidence of $conf^{(c)}$, if $conf^{(c)}\geq \delta_k^{(c)}$, it will be reckon as a reliable prediction. Otherwise, it will be abandoned. This process can be formatted as:

\begin{equation}
    \hat{y}_l^{(\hat{c})}=\left\{
\begin{array}{cc}
1, & if\enspace conf^{(\hat{c})}=\mathop{argmax}\limits_{c}\thinspace conf^{(c)} \\
& and\enspace conf^{(\hat{c})}\geq \delta_k^{(\hat{c})}, \\
 & \\
0, & otherwise,
\end{array}
\right.
\end{equation}

\noindent where $\hat{y}_l^{(\hat{c})}$ is the one-hot label derived from the $l$-th prediction result, $conf^{(c)}$ is the confidence for class $c$.

These pseudo labels, denoted as $\{\hat{y}_l^T,\hat{b}_l^T\}_{l=1}^m,$ are used to train the student model on the target domain with loss $L_{unsup}$, which is the same to $L_{sup}$ in format:

\begin{equation}
    L_{unsup}=\lambda_1L_{bbox}(\hat{b}_i, \hat{b}_i^T)+\lambda_2L_{obj}(\hat{y}_i,\hat{y}_i^T)+\lambda_3L_{cls}(\hat{y}_i,\hat{y}_i^T).
\end{equation}

\subsubsection{Update Student Model and Teacher Model}

On both domains, the student model updates its parameters through backpropagation. In parallel, the teacher model updates its parameters via EMA after the student model's update:

\begin{equation}
    w_k^{(t)}=\alpha w_{k-1}^{(t)}+(1-\alpha)w_k^{(s)},
\end{equation}

\noindent where $w_k^{(t)}$ and $w_k^{(s)}$ represent the parameters of the teacher and student models at step $k$ respectively. The teacher model's parameters are essentially a weighted sum of past student model parameters, determined as follows:

\begin{equation}
    \begin{array}{rcl}
w_k^{(t)} & = & \alpha^k w_0^{(s)}+\sum_{i=1}^k \alpha^{k-i}(1-\alpha)w_i^{(s)} \\
 & = & \alpha^k w_0^{(s)}+\alpha^{k-1}(1-\alpha)w_1^{(s)}+\cdots+(1-\alpha)w_k^{(s)},
\end{array}
\end{equation}

\noindent where $w_0^{(t)}=w_0^{(s)}$.

\subsection{Adversarial Learning}\label{sec:al}

The challenge in UDA lies in bridging the gap between two domains with differing data distributions. This discrepancy results in distinct representations of the same object in the latent feature space, a misalignment we seek to mitigate. To accomplish this, our approach employs image-level alignment, instance-level alignment, and consensus regulation.

\subsubsection{Image-level Alignment} The features extracted by the network will be fed into image-level discriminator to predict which domain this image in. The discriminator tries to distinguish where this image comes from while the network endeavors to deceive the discriminator. The loss can be expressed as follows:

\begin{equation}
    L_{img}=-\lambda_{d}\sum_s l(D_{img}(f_i^s),d_i),
\end{equation}

\noindent in which $f_i^s$ represents feature of the $i$-th image, $s$ indicates the scale of the feature, $D$ is the image-level discriminator. For instance, YOLOv5 will output features in three different scales. $d_i$ is its domain label. $l$ is the loss function, and $\lambda_{d}$ is the weight assigned to this discriminator.

\subsubsection{Targeted Instance-level Alignment} It is also crucial for the network to extract domain-invariant feature on instance level. Additionally, we emphasize foreground pixels over background pixels, as this distinction aids in object detection. In two-stage architectures, it is done by RPN, using the regions RPN proposed to mark where objects might exist. Then a FCN is adopted to predict every pixel's domain label. However, one-stage detectors have no module similar to RPN, which makes instance-level alignment more challenging. Previous methods just omit location information of the objects, performing instance-level alignment on every pixel equally, or add an additional module to predict location. Our innovation involves extracting location information from a teacher model's predictions.

After the two-step selection, pseudo labels are acquired and denoted as $\{\hat{y}^T_l,\hat{b}^T_l\}_{l=1}^m$. The pseudo bounding boxes $\hat{b}^T_l=(x_1,y_1,x_2,y_2,conf,c_1,\cdots,c_n)$ consists the confidence $conf$, which shows whether there exists an object on every pixel, which can effectively serve as a saliency matrix indicating location information. We use $m^s$ to represent the saliency matrix, where $s$ denoting its scale. $m^s$ is used to reweigh the feature map to boost stimulation on objects' location. The reweighing process can be formulated as:

\begin{equation}
    \bar{f}_i^s=(m^s\otimes f_i^s)\oplus f_i^s,
\end{equation}

\noindent in which $\otimes$ means Hadamard product, $\oplus$ represents pixel-wise addition. The targeted instance-level loss can be written as:

\begin{equation}
    L_{ins}=-\lambda_{d} \sum_s\sum_{u,v}l(D_{ins}(\bar{f}_i^s)^{(u,v)},d_i^{(u,v)}),
\end{equation}

\noindent where $(u,v)$ denotes position on the feature map, $D_{ins}$ is the instance-level discriminator, $d_i^{(u,v)}$ is the domain label for position $(u,v)$, $\lambda_{d}$ is the corresponding weight. During backpropagation, instance-level discriminators impose additional penalties on the target-existing position, which shares the same idea with the vanilla instance-level alignment.

\begin{figure}
    \centering
    \includegraphics[width=0.9\textwidth]{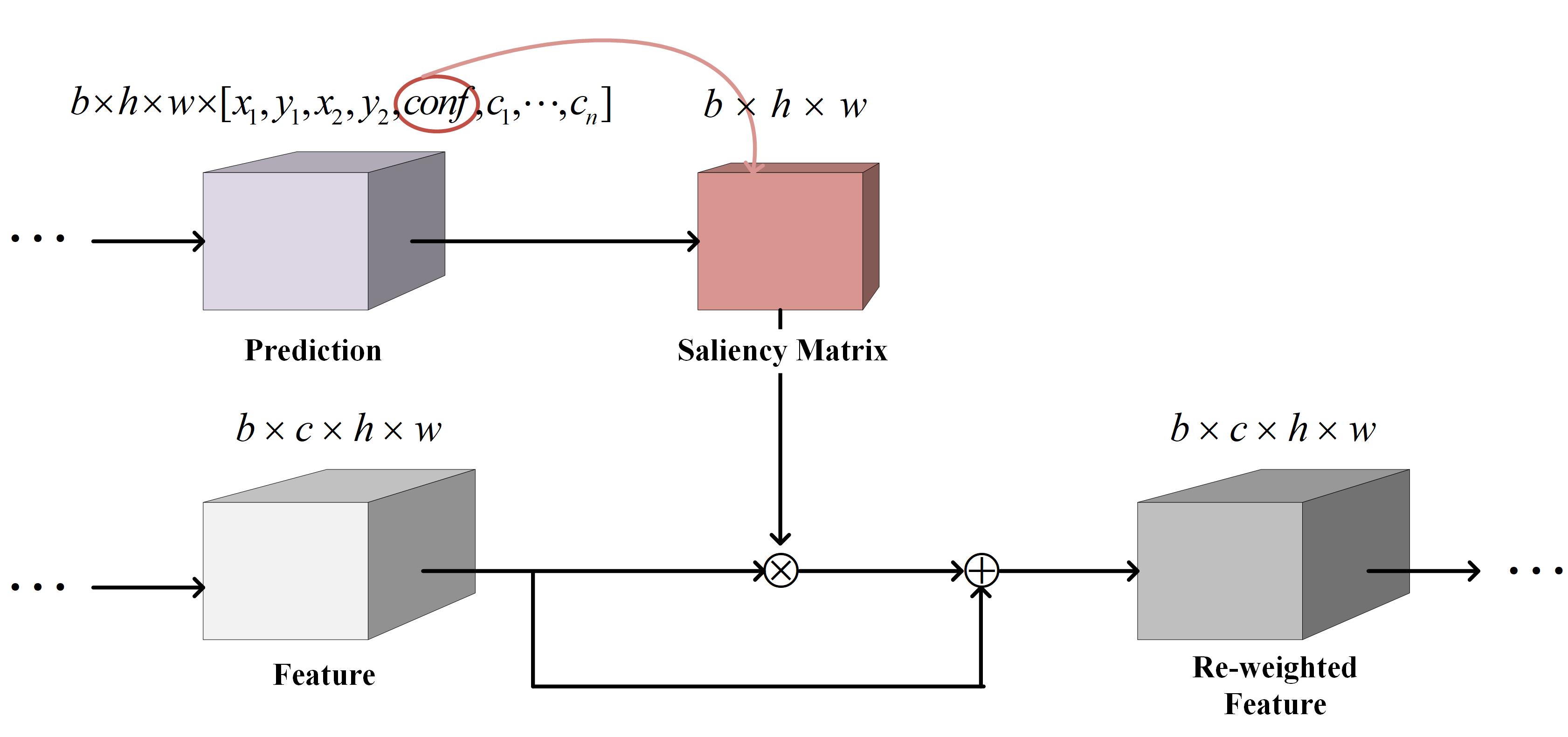}
    \caption{The procedure of how the saliency matrix works. Saliency matrix is derived from pseudo labels, and is used to compute the weighted features through dot product with the extracted features. Features corresponding to the foreground are enhanced, while features related to the background are suppressed.}
    \label{fig:attention}
\end{figure}

In supervised scenarios, where images from the source domain have corresponding labels, the saliency matrix is directly generated from the labels, highlighting the regions where objects are present and suppressing the rest.

\subsubsection{Consensus Regulation} To encourage consistency between the two discriminators, we introduce consensus regulation. This regularization penalizes differences in their outputs and ensures alignment, which is defined as:

\begin{equation}
    L_{con} = -\lambda_{d} \sum_s \sum_{u,v} l(D_{ins}(\bar{f}_i^s)^{(u,v)}, D_{img}(f_i^s)).
\end{equation}

\subsection{Summarize}

In the training process of our proposed method, the total loss, denoted as $L$, is composed of several key components:

\begin{equation}
    L=L_{sup}+L_{unsup}+\lambda_{d}(L_{img}+L_{ins}+L_{con}),
\end{equation}

\noindent where $\lambda_{d}$ is hyper-parameter to control weight of adversarial learning. $L_{sup}$ and $L_{unsup}$ come into play when processing data from the source and target domains, respectively.  All of these loss terms collectively contribute to updating the student model during training, while the teacher model is updated via EMA.

\section{Experiments}

\subsection{Implementation Details}

% YOLOv5-L is employed as the detection model in our VT framework as well as all other compared models. All models are pretrained on source domain. 

The experiments are conducted using two paradigms: YOLO and Faster R-CNN, with a primary focus on the YOLO paradigm. The rationale behind choosing YOLO is closely tied to our contributions. While YOLO is frequently chosen as a backbone in practice due to its speed advantages, the absence of RPN in one-stage detectors presents challenges for aligning instance features, limiting its applications. One of our main contributions is the integration of location-aware instance-level alignment with one-stage detectors, without relying on proposals from RPN. 

Following the implementation of YOLO series, images are resized to 640 $\times$ 640. For the hyper parameters, following literature \cite{yu-jheliCrossDomainAdaptiveTeacher2022}, we set $\lambda_{d}=0.1$ for all datasets. Regarding confidence thresholds, we use $\delta_t=0.2$ to filter out objects with low confidence scores. Additionally, $\delta_0^{(c)}$ is initialized to 0.8 for all object classes but is subject to adaptation during training. To stabilize training, we incorporate Exponential Moving Averages (EMA) with weights $\alpha=0.9996$ and $\alpha_t=0.9999$. VT is trained with an initial learning rate of $1\times 10^{-4}$, which is linearly decreased to $1\times 10^{-6}$ over 50 epochs. We employ Stochastic Gradient Descent (SGD) to optimize the model's parameters. 

Data augmentation plays a crucial role in our experiments. Random horizontal flip acts as weak augmentation for both model. Color jitter, Gaussian blur and random erase are used as strong augmentation for the student model. All experiment is conducted on a NVIDIA RTX 3090 with batch size of 16 and implemented with PyTorch.

As for employing Faster R-CNN detector, we follow the setting of AT\cite{yu-jheliCrossDomainAdaptiveTeacher2022}, VGG16 is selected as the backbone. For hyperparemeters, we adapt the same parameters in YOLO paradigm. All models are trained for 100,000 iterations.

\subsection{Datasets}

Our experiments involve four datasets, each serving a specific purpose in evaluating unsupervised domain adaptation:

\textbf{Cityscapes.} Cityscapes\cite{Cordts2016Cityscapes} is a large-scale dataset that contains a diverse set of stereo video sequences recorded in street scenes from 50 different cities. It provides high-quality pixel-level annotations for 5,000 frames. For our experiments, we use 2975 images for training and 500 images for validation.

\textbf{Foggy Cityscapes.} Foggy Cityscapes\cite{SDV18} is based on Cityscapes with synthesized fog in different level. 
Thus it shares the same structure as Cityscapes. This dataset allows us to assess our model's ability to adapt to different weather conditions. We perform domain adaptation from Cityscapes to Foggy Cityscapes.

\textbf{KITTI.} KITTI\cite{Geiger2012CVPR} consists of hours of traffic scenarios recorded with a variety of sensor modalities, including high-resolution RGB, grayscale stereo cameras, and a 3D laser scanner. We use 4717 images for training and 520 images for validation. In this case, we conduct domain adaptation from KITTI to Cityscapes, focusing on the impact of camera devices and perspective. We select shared classes (person and car) and exclude others.

\textbf{Sim10k.} Sim10k\cite{johnson2017driving} is a synthetic dataset containing 10,000 images rendered from the video game Grand Theft Auto V (GTA5). We split the dataset into 6136 training images and 684 validation images. Here, we assess the ability to adapt from Sim10k to Cityscapes, bridging the domain gap between real and synthesized images. Sim10k exhibits extreme class imbalance, making it unsuitable to initial UDA. Following the literature\cite{cheng-chunhsuEveryPixelMatters2020}, only class car is considered.

% Table generated by Excel2LaTeX from sheet 'Sheet1'
\begin{table*}[htbp]
  \centering
  \caption{The result and comparison on adversarial weather adaptation, testing for \textbf{Cityscapes} to \textbf{Foggy Cityscapes} adaptation under YOLO paradigm. Mean average precision on all classes is presented.}
  \scalebox{0.9}{
    \begin{tabular}{cccccccccc}
    \hline
    \textbf{Model} & \multicolumn{9}{c}{\textbf{mAP@0.5}} \\
     (YOLOv5-L)     & \textbf{Car} & \textbf{Bicycle} & \textbf{Person} & \textbf{Rider} & \textbf{Mcycle} & \textbf{Bus} & \textbf{Truck} & \textbf{Train} & \textbf{All} \\
    \hline
    Baseline & 61.2  & 38.0    & 46.6  & 48.0    & 33.8  & 45.6  & 27.6  & 39.3  & 42.5 \\
    MSDAYOLO & 60.9  & 36.8  & 44.3  & 43.6  & 29.5  & 48.0 & 30.8  & 36.4  & 41.3 \\
    Integrated YOLO & 64.9  & 38.7  & 48.4  & 48.4  & 32.9  & 51.8  & 30.3  & 40.4  & 44.5 \\
    DAYOLO & 65.2  & 39.8  & 48.3  & 48.1  & 30.6  & 53.7  & 33.8  & 45.1  & 45.6 \\
    EPM   & 65.9  & 39.6  & 48.0    & 49.1  & 33.3  & 49.9  & 31.2  & 39.7  & 44.6 \\
    \hline
    AT    & 62.1  & 39.4  & 45.9 & 45.0  & 31.1  & 50.1 & 29.5 & 43.0  & 43.3 \\
    SSDA-YOLO    & \textbf{69.8}  & \textbf{46.0}  & \textbf{50.1} & 51.2 & 38.6  & 59.7  & 34.9  & 47.9 & 49.7 \\
    CMT    & 61.2  & 40.8  & 48.0  & 49.4  & 33.1  & 51.9  & 33.4 & 41.2  & 44.9 \\
    VT  & 65.9 & 44.1 & 49.2 & \textbf{51.5} & \textbf{38.7} & \textbf{63.0}  & \textbf{40.0}  & \textbf{56.3} & \textbf{51.1} \\
    \hline
    Oracle & 69.9  & 40.0  & 50.3  & 48.5  & 34.2  & 55.2  & 37.3  & 35.9  & 46.4 \\
    \hline
    \textbf{Model} & \multicolumn{9}{c}{\textbf{mAP@0.5:0.05:0.95}} \\
     (YOLOv5-L)     & \textbf{Car} & \textbf{Bicycle} & \textbf{Person} & \textbf{Rider} & \textbf{Mcycle} & \textbf{Bus} & \textbf{Truck} & \textbf{Train} & \textbf{All} \\
    \hline
    Baseline & 42.9  & 19.8  & 25.7  & 26.7  & 15.4  & 35.5  & 19.8  & 14.3  & 25.0 \\
    MSDAYOLO & 39.8  & 18.0    & 23.4  & 22.4  & 13.7  & 33.6  & 20.4  & 12.4  & 23.0 \\
    Integrated YOLO & 43.9  & 19.6  & 26.2  & 27.2  & 15.2  & 41.0    & 22.0    & 14.9  & 26.2 \\
    DAYOLO & 43.8  & 19.8  & 25.8  & 26.2  & 14.6  & 41.9  & 25.2  & 17.2  & 26.8 \\
    EPM   & 44.7  & 20.3  & 26.2  & 27.7  & 13.9  & 39.7  & 23.2  & 13.9  & 26.2 \\
    \hline
    AT    & 42.1  & 19.6  & 25.2  & 25.1  & 15.1  & 39.3  & 22.2 & 16.6  & 25.6 \\
    SSDA-YOLO    & \textbf{45.3}  & 17.9  & 26.1  & 27.0  & 10.0  & \textbf{52.7}  & 26.6 & 16.6  & 27.8 \\
    CMT    & 41.4  & 21.3  & 26.2  & 28.8  & 15.7  & 41.2  & 24.4 & 15.4  & 26.8 \\
    VT  & 44.7 & \textbf{22.4} & \textbf{26.6} & \textbf{30.3} & \textbf{18.3} & 49.4 & \textbf{29.6}  & \textbf{23.1} & \textbf{30.5} \\
    \hline
    Oracle & 47.4  & 20.0  & 27.1  & 27.1  & 13.2  & 41.6  & 25.8  & 16.3  & 27.3 \\
    \hline
    \end{tabular}%
    }
  \label{tab:awa}%
\end{table*}%

% Table generated by Excel2LaTeX from sheet 'Sheet1'
\begin{table*}[htbp]
  \centering
  \caption{The result and comparison on adversarial weather adaptation, testing for \textbf{Cityscapes} to \textbf{Foggy Cityscapes} adaptation under Faster R-CNN paradigm. Mean average precision on all classes is presented.}
  \scalebox{0.9}{
    \begin{tabular}{cccccccccc}
    \hline
    \textbf{Model} & \multicolumn{9}{c}{\textbf{mAP@0.5}} \\
     (F-RCNN)     & \textbf{Car} & \textbf{Bicycle} & \textbf{Person} & \textbf{Rider} & \textbf{Mcycle} & \textbf{Bus} & \textbf{Truck} & \textbf{Train} & \textbf{All} \\
    \hline
    Baseline & 39.6  & 31.9    & 29.0  & 37.2    & 16.9  & 20.1  & 8.1  & 5.2  & 23.5 \\
    DA-Faster & 40.5  & 27.1  & 25.0  & 31.0  & 20.0  & 35.3 & 22.1  & 20.2  & 27.6 \\
    SCDA & 48.5  & 33.6  & 33.5  & 38.0  & 28.0  & 39.0 & 26.5  & 23.3  & 33.8 \\
    DM & 44.3  & 32.2  & 30.8  & 40.5  & 28.4  & 38.4 & 27.2  & 34.5  & 34.6 \\
    \hline
    MTOR & 44.0  & 35.6  & 30.6  & 41.1  & 28.3  & 38.6 & 21.9  & 40.6  & 35.1 \\
    CRDA & 49.2  & 34.6  & 32.9  & 43.8  & 30.3  & 45.1 & 27.2  & 36.4  & 37.4 \\
    UMT & 48.6  & 37.3  & 33.0  & 46.7  & 33.4  & 56.5 & \textbf{34.1}  & 46.8  & 41.7 \\
    AT    & 63.8  & \textbf{53.0}  & 43.2 & 52.7  & 34.4  & 58.4 & 32.4 & 34.5  & 46.6 \\
    CMT    & 62.5  & 52.3  & 42.3 & 52.9  & 38.6  & \textbf{58.6} & 31.7 & 40.9  & 47.5 \\
    VT    & \textbf{63.9}  & 52.3  & \textbf{44.0} & \textbf{53.3} & \textbf{38.8}  & 56.8  & 31.0  & \textbf{48.4} & \textbf{48.6} \\
    \hline
    Oracle & 61.3  & 40.7  & 43.1  & 49.8  & 32.5  & 50.3  & 28.6  & 35.1  & 42.7 \\
    \hline
    \end{tabular}%
    }
  \label{tab:awa-f}%
\end{table*}%

\subsection{Results}

In this section, we present the results of our research, comparing our method (denoted as VT) with existing state-of-the-art methods. For YOLO paradigm, we compared our method with MSDAYOLO\cite{mazinhnewaMultiscaleDomainAdaptive2021}, Integrated MSDAYOLO\cite{mazinhnewaIntegratedMultiscaleDomain2023a} (denoted as Integrated YOLO), DAYOLO\cite{shizhaozhangDomainAdaptiveYOLO2021}, EveryPixelMatter (EPM)\cite{cheng-chunhsuEveryPixelMatters2020}, Adaptive Teacher (AT)\cite{yu-jheliCrossDomainAdaptiveTeacher2022}, SSDA-YOLO\cite{zhou2023ssda} and CMT\cite{cao2023contrastive}. It is notable that MSDAYOLO, Integrated YOLO, DAYOLO and EPM only use adversarial-based methods, while AT, SSDA-YOLO, CMT and our method employ a teacher-student architecture. YOLOv5-L is utilized for every method. Evaluation is based on two key metrics: \textit{mAP@0.5} and \textit{mAP@0.5:0.05:0.95}. In our study on the Faster R-CNN paradigm, we conducted an analysis with several methods, including DA Faster R-CNN\cite{chen2018domain} (denoted as DA-Faster), SCDA\cite{zhu2019adapting}, Diversify and Match\cite{kim2019diversify} (denoted as DM), MTOR\cite{cai2019exploring}, CRDA\cite{xu2020exploring}, UMT\cite{deng2021unbiased} and AT\cite{yu-jheliCrossDomainAdaptiveTeacher2022}. Notably, MTOR, CRDA, UMT and AT adopt a teacher-student framework. Evaluation is based on \textit{mAP@0.5}. We introduce an ``Oracle'' result, representing a scenario where the model is trained and tested on the target domain.

\begin{figure*}[htbp]
    \centering
    \subcaptionbox{$\delta=0.7$\label{fig: 07}}
	{\includegraphics[width=0.22\linewidth]{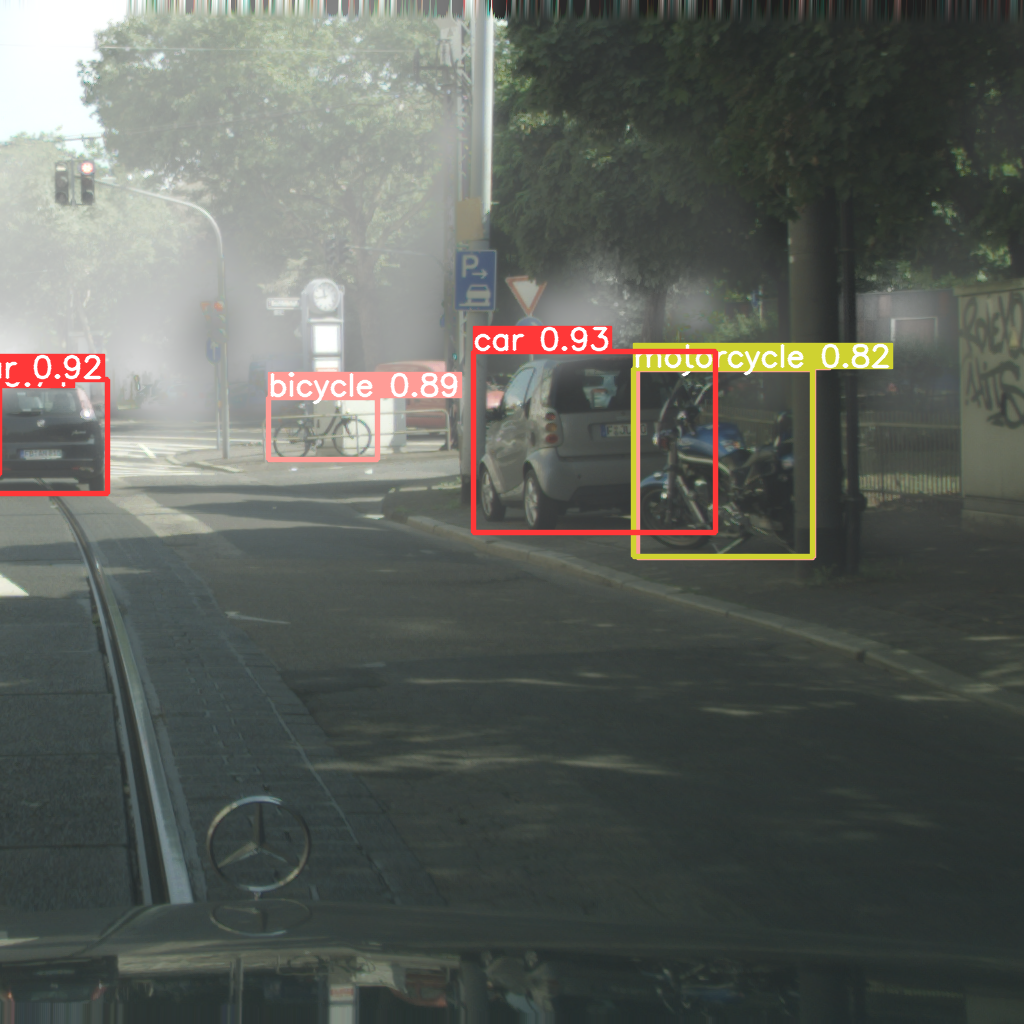}}
    \quad
    \subcaptionbox{$\delta=0.8$\label{fig: 08}}
	{\includegraphics[width=0.22\linewidth]{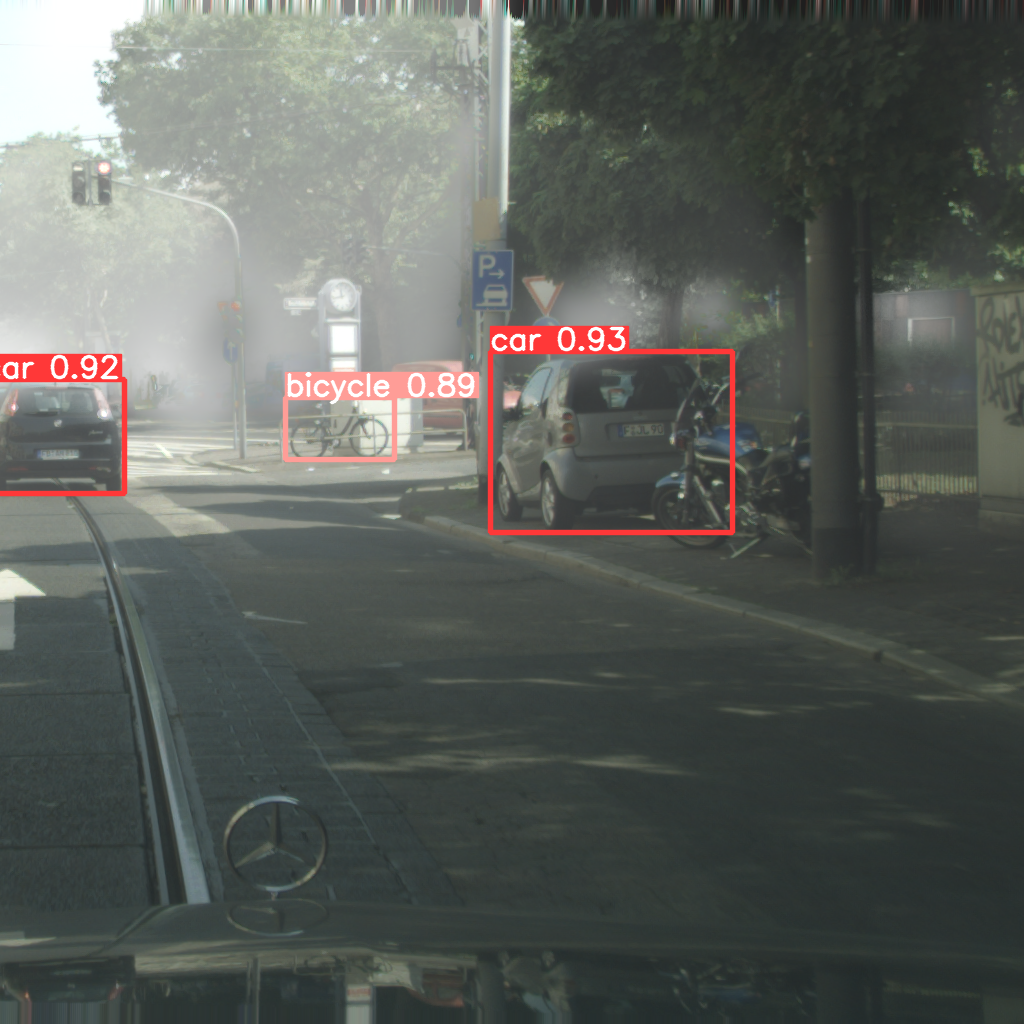}}
    \quad
    \subcaptionbox{$\delta=0.9$\label{fig: 09}}
	{\includegraphics[width=0.22\linewidth]{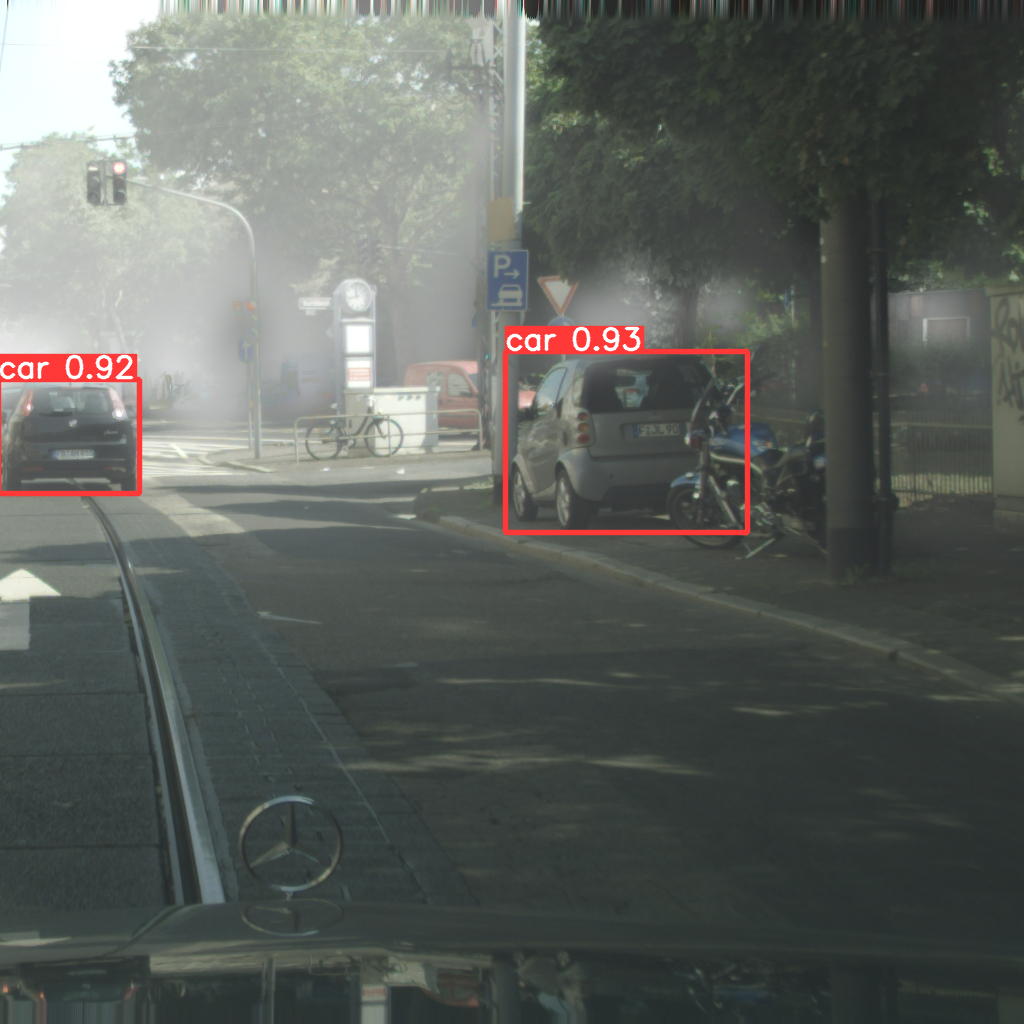}}
    \quad
    \subcaptionbox{CAPS\label{fig: caps}}
	{\includegraphics[width=0.22\linewidth]{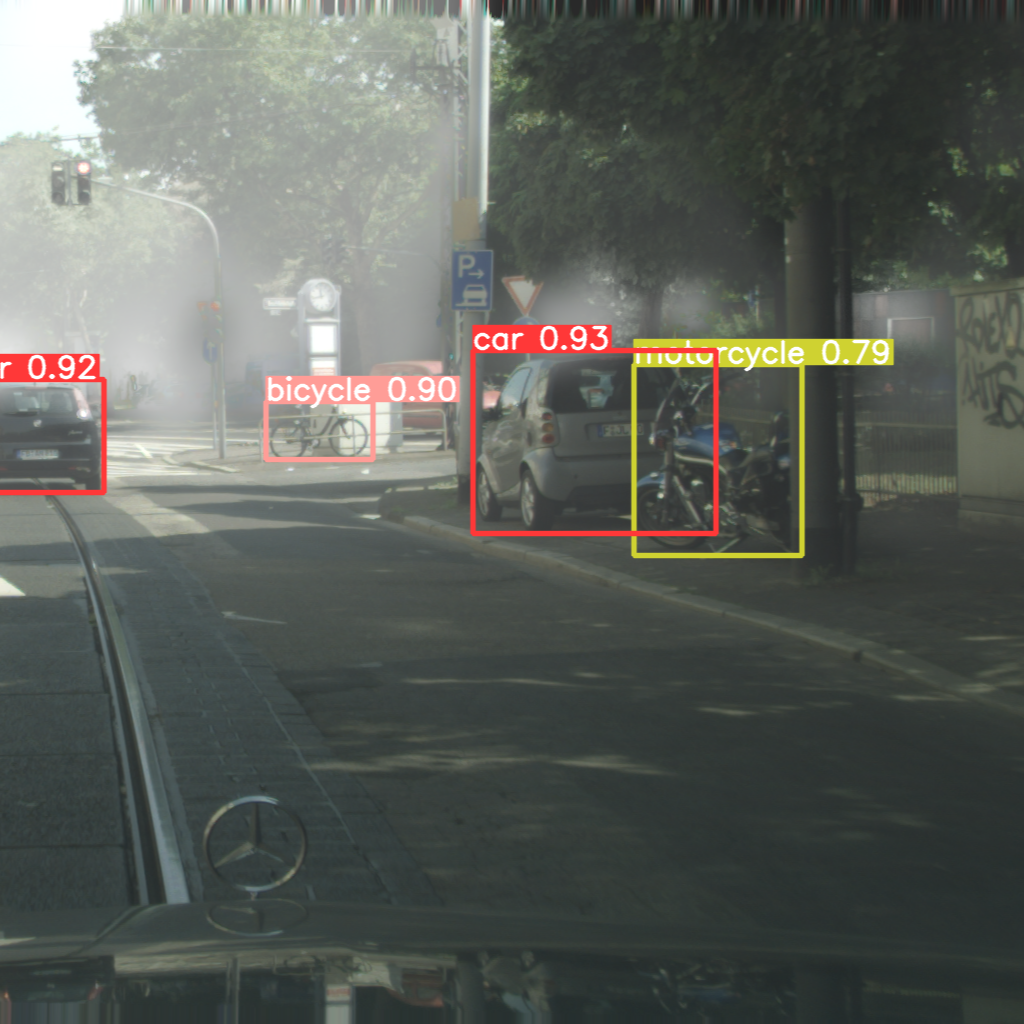}}
    \caption{The comparison of setting a static global threshold and employing CAPS mechanism. The bounding boxes are the pseudo labels generated by the teacher model. $\delta$ is the global threshold for all classes.}
    \label{fig:diff}
\end{figure*}

\textbf{Adverse Weather Adaptation.} Table \ref{tab:awa} and \ref{tab:awa-f} showcase the results of our Adverse Weather Adaptation experiments. In these challenging conditions, VT achieves state-of-the-art (SOTA) performance. Specifically, VT reaches an impressive 51.1\% \textit{mAP@0.5} and 30.5\% \textit{mAP@0.5:0.05:0.95} when using YOLOv5-L as the backbone within the YOLO paradigm. Furthermore, with VGG16 as the backbone in the Faster R-CNN paradigm, VT attains 48.6\% \textit{mAP@0.5}. The introduction of a class-aware threshold has been proven instrumental, enabling our teacher model to accurately capture correctly classified objects and generate highly informative pseudo labels, thereby contributing to a more accurate result. In Figure \ref{fig:diff}, we present a comparative analysis between setting a static global threshold and implementing the CAPS mechanism. A relatively low threshold can result in the generation of incorrect pseudo labels. In Figure \ref{fig: 07}, a motorcycle is erroneously labeled as both a motorcycle and a bicycle. Conversely, setting a high threshold may lead to the neglect of real targets, as seen in Figure \ref{fig: 08} and Figure \ref{fig: 09}, where the motorcycle is not labeled. However, the proposed CAPS mechanism can perform a class-aware selection and generate more reliable pseudo labels.

\textbf{Cross-camera Adaptation.} Our Cross-camera Adaptation experiments, as presented in the left part of Table \ref{tab:aca}, demonstrate the superiority of VT framework over competing approaches. As discussed earlier, the disparity in difficulty when detecting cars and people between the two datasets poses a significant challenge. Utilizing a fixed threshold for pseudo label selection, as employed by AT, results in either neglecting true objects or incorporating erroneous information. Leveraging an adaptable threshold, our approach achieves a higher level of precision.

% Table generated by Excel2LaTeX from sheet 'Sheet1'
\begin{table}[htbp]
  \centering
  \caption{The left part shows the result and comparison on cross-camera adaptation, testing for \textbf{KITTI} to \textbf{Cityscapes} adaptation. Only class car and class person are used. The right part shows the result and comparison on cross-camera adaptation, testing for \textbf{Sim10k} to \textbf{Cityscapes} adaptation. Only class car is used in the experiment. Mean average precision on all classes is presented.}
  \scalebox{0.95}{
    \begin{tabular}{ccccccc|cc}
    \hline
    %\multicolumn{7}{c}{\textbf{Cross-camera}} & \multicolumn{2}{c}{\textbf{Style}} \\
    \textbf{Model} & \multicolumn{3}{c}{\textbf{mAP@.5}} & \multicolumn{3}{c|} 
    {\textbf{mAP@.5:.95}} & \textbf{mAP@.5} & \textbf{mAP@.5:.95}\\
     (YOLOv5-L)     & \textbf{Car} & \textbf{Person} & \textbf{All} & \textbf{Car} & \textbf{Person} & \textbf{All} & \textbf{Car} & \textbf{Car}\\
    \hline
    Baseline & 56.2  & 36.9  & 46.5  & 32.9  & 18.0    & 25.5 & 58.7 & 36.9\\
    MSDAYOLO & 56.5  & 37.0    & 46.7  & 32.3  & 17.7  & 25.0 & 63.0 & 35.6\\
    Integrated YOLO & 56.4  & 36.9  & 46.6  & 32.6  & 17.7  & 25.1 & 64.2 & 37.9\\
    DAYOLO & 57.4  & 37.4  & 47.4  & 34.1  & 18.3  & 26.2 & 63.9 & 36.9\\
    EPM   & 58.1  & 38.5  & 48.3  & 34.2  & 18.8  & 26.5 & 64.7 & 37.6\\
    \hline
    AT    & 59.0 & 38.3  & 48.7  & 34.8 & 19.3  & 26.8 & 56.4 & 33.9\\
    SSDA-YOLO    & 59.6 & 32.2  & 45.9  & 33.0 & 12.1  & 22.5 & 60.1 & 33.7\\
    CMT    & 53.2 & 40.4  & 46.8  & 29.8 & 20.0  & 24.9 & 60.3 & 37.8\\
    VT  & \textbf{61.9}  & \textbf{43.2} & \textbf{52.6} & \textbf{37.3}  & \textbf{21.9} & \textbf{29.6} & \textbf{65.3} & \textbf{38.3} \\
    \hline
    Oracle  & 72.2  & 51.7  & 62.0  & 48.4  & 27.7  & 38.0 & 73.9 & 50.5\\
    \hline
    \end{tabular}%
    }
  \label{tab:aca}%
\end{table}%

\textbf{Synthetic to Real Adaptation.} The results for Synthetic to Real Adaptation are presented in the right part of Table \ref{tab:aca}. Since Sim10k is captured in a video game, the images is lacking in realistic details, which makes it hard to transfer. However, VT can leverage saliency matrices derived from labels to focus on object-related features, extracting more universal features for detection. As a result, VT secures best accuracy among all the methods.

% Table generated by Excel2LaTeX from sheet 'Sheet1'
\begin{table*}[htbp]
  \centering
  \caption{The ablation study on our VT. IA means Instance-level Alignment, PIA means Position-aware Instance-level Alignment, which combines IA with Saliency Matrices, CAPS means Class-aware Adaptive Pseudo-label Selection.}
  \scalebox{0.95}{
    \begin{tabular}{cccccccccc}
    \hline
    \textbf{Model} & \multicolumn{9}{c}{\textbf{mAP@0.5}} \\
     (YOLOv5-L)     & \textbf{Car} & \textbf{Bicycle} & \textbf{Person} & \textbf{Rider} & \textbf{Mcycle} & \textbf{Bus} & \textbf{Truck} & \textbf{Train} & \textbf{All} \\
    \hline
    Baseline    & 62.1  & 39.4  & 45.9 & 45.0  & 31.1  & 50.1 & 29.5 & 43.0  & 43.3 \\
    IA & 63.6  & 43.3  & 48.7  & 50.9  & \textbf{38.9}  & 59.7  & 34.4 & 41.3  & 47.6 \\
    TIA & 64.6  & 42.3  & 48.1  & 48.3  & 36.6  & 60.6  & 39.5  & 49.4  & 48.7 \\
    CAPS & 63.9  & 42.1 & 48.3 & 50.8 & 37.3  & 61.3    & 39.5 & 51.2  & 49.3 \\
    TIA \& CAPS    & \textbf{65.9} & \textbf{44.1}  & \textbf{49.2}  & \textbf{51.5} & 38.7 & \textbf{63.0} & \textbf{40.0}  & \textbf{56.3} & \textbf{51.1} \\
    \hline
    \textbf{Model} & \multicolumn{9}{c}{\textbf{mAP@0.5:0.05:0.95}} \\
     (YOLOv5-L)     & \textbf{Car} & \textbf{Bicycle} & \textbf{Person} & \textbf{Rider} & \textbf{Mcycle} & \textbf{Bus} & \textbf{Truck} & \textbf{Train} & \textbf{All} \\
    \hline
    Baseline    & 42.1  & 19.6  & 25.2  & 25.1  & 15.1  & 39.3  & 22.2 & 16.6  & 25.6 \\
    IA & 41.7  & 22.1  & 25.7  & 29.3  & 17.3  & 44.7  & 23.7    & 18.1  & 27.8 \\
    TIA & 44.0  & 22.0  & 26.6 & 28.8    & 17.5  & 47.3 & 28.5  & 22.8 & 29.7 \\
    CAPS & 44.6  & 21.8  & \textbf{26.9}  & 29.9  & 17.3  & 48.0  & 28.7 & \textbf{23.6}  & 30.1 \\
    TIA \& CAPS    & \textbf{44.7} & \textbf{22.4} & 26.6  & \textbf{30.3} & \textbf{18.3} & \textbf{49.4} & \textbf{29.6} & 23.1  & \textbf{30.5} \\
    \hline
    \end{tabular}%
    }
  \label{tab:ablation}%
\end{table*}%

\subsection{Ablation Studies}

In this section, we conduct ablation studies to assess the effectiveness of two key components: the saliency matrix used in the instance-level alignment stage and the proposed CAPS. The results of these ablation studies are summarized in Table \ref{tab:ablation}. Specifically, we examine the following variations of our approach: AT (as the baseline), AT with instance-level alignment (denotes as IA), AT with targeted instance-level alignment (denotes as TIA), AT with CAPS, and AT with both TIA and CAPS (i.e. VT). These experiments are conducted in an adversarial weather adaptation setting, where domain adaptation is performed from the Cityscapes dataset to Foggy Cityscapes.

We report the results using the same evaluation metrics. The results indicate improvements brought about by both the saliency matrix and the CAPS mechanism, particularly for challenging classes such as ``train'' because of its scarcity.

The saliency matrix plays a crucial role in enabling the feature extractor to direct its focus specifically towards the spatial locations of objects and guiding it to extract more discriminative features, thereby improving the overall performance. Meanwhile, the CAPS mechanism contributes by executing a class-aware selection and generating pseudo labels with heightened reliability. In cases where certain classes pose challenges for detection, CAPS considers predictions with lower confidence levels, ensuring adequate supervision for these classes. Otherwise, there may be no pseudo label for those classes. Conversely, it employs a discerning approach for classes that are easily detected, filtering out potential noise. From another perspective, CAPS can prevent the model from overfitting on easily detected classes. If there is a continuously absence of supervision for hard classes, i.e. no pseudo labels for hard classes are generated, the model might exclusively prioritize easy classes, potentially leading to overfitting.

Utilizing either the saliency matrix or the CAPS mechanism alone leads to performance improvements, demonstrating their individual effectiveness. Moreover, combining both components results in even better overall performance.

\section{Conclusion}

This paper introduces the Versatile Teacher, a novel teacher-student framework to address the domain adaptation challenge in object detection task. In fact, our framework is module-agnostic, allowing compatibility with both one- and two-stage detectors. Our contribution lies in the development of the Class-aware Adaptively Selection (CAPS) mechanism, which significantly improves the reliability of pseudo labels generated by the teacher model. Moreover, we have harnessed these pseudo labels as saliency matrices, guiding instance-level alignment to emphasize object-region features. The experiments conducted on three benchmark datasets demonstrated the effectiveness of our approach in cross-domain adaptation, showcasing promising results. Furthermore, an extended ablation study shows the effectiveness of each proposed module.

\section*{Acknowledgement}

This research was supported by National Natural Science Foundation of China (Grant No. 42071339).

%% The Appendices part is started with the command \appendix;
%% appendix sections are then done as normal sections
% \appendix
% \section{Example Appendix Section}
% \label{app1}

% Appendix text.

%% For citations use: 
%%       \cite{<label>} ==> [1]

%%
% Example citation, See \cite{lamport94}.

%% If you have bib database file and want bibtex to generate the
%% bibitems, please use
%%
%%  \bibliographystyle{elsarticle-num} 
%%  \bibliography{<your bibdatabase>}

%% else use the following coding to input the bibitems directly in the
%% TeX file.

%% Refer following link for more details about bibliography and citations.
%% https://en.wikibooks.org/wiki/LaTeX/Bibliography_Management

\bibliographystyle{elsarticle-num}
\bibliography{refs}
\end{document}